\documentclass[runningheads]{llncs}
\usepackage[T1]{fontenc}
\usepackage{graphicx,verbatim}
\usepackage[acronym,nomain]{glossaries}
\makeglossaries
\newacronym{sam}{SAM}{Segment anything model}
 
\newacronym{vit}{ViT}{Vision Transformer}
\newacronym{mlp}{MLP}{Multilayer Perceptron}
\newacronym{mae}{MAE}{Masked Auto-Encoder}

\newacronym{kd}{KD}{knowledge distillation}
\newacronym{dcrf}{DCRF}{Dense Conditional Random Field}
\newacronym{iou}{IoU}{ Intersection over Union }
\newacronym{top}{TOP}{ transient object presence}
  
\newacronym{sbert}{Slot-BERT}{}
\newacronym{bert}{BERT}{}
\newacronym{vae}{VAE}{Variational Auto-encoder}
\newacronym{sa}{SA}{Slot Attention}
\newacronym{pca}{PCA}{Principal Component Analysis }
\newacronym{nlp}{NLP}{Natural Language Processing}
 
\newacronym{tst}{TST}{Temporal Slot Transformer}
 
\newacronym{hd}{HD}{Hausdorff distance}
\newacronym{llm}{LLM}{Large language Models}

\newacronym{dtst}{DTST}{Dynamic Temporal Slot Transformer}

\usepackage{amsmath,amssymb,amsfonts}
\usepackage{graphicx}
\usepackage{textcomp}
\def\BibTeX{{\rm B\kern-.05em{\sc i\kern-.025em b}\kern-.08em
    T\kern-.1667em\lower.7ex\hbox{E}\kern-.125emX}}

\usepackage{xargs}                      % Use more than one optional parameter in a new commands
\usepackage[pdftex,dvipsnames]{xcolor}  % Coloured text etc.
\usepackage[colorinlistoftodos,prependcaption,textsize=small]{todonotes}
\newcommandx{\unsure}[2][1=]{\todo[linecolor=red,backgroundcolor=red!25,bordercolor=red,#1]{#2}}
\newcommandx{\change}[2][1=]{\todo[linecolor=blue,backgroundcolor=blue!25,bordercolor=blue,#1]{#2}}
\newcommandx{\info}[2][1=]{\todo[linecolor=OliveGreen,backgroundcolor=OliveGreen!25,bordercolor=OliveGreen,#1]{#2}}
\newcommandx{\improvement}[2][1=]{\todo[linecolor=Plum,backgroundcolor=Plum!25,bordercolor=Plum,#1]{#2}}
\newcommandx{\thiswillnotshow}[2][1=]{\todo[disable,#1]{#2}}

\usepackage{hyperref}
\hypersetup{
    colorlinks=true,
    linkcolor=blue, 
    filecolor=blue,      
    urlcolor=blue,
    citecolor=blue, % Set the color of citations
}

\usepackage{multirow}
\usepackage{tabularx}
\usepackage{hhline}
\usepackage{array}
\newcolumntype{P}[1]{>{\centering\arraybackslash}p{#1}}
\newcolumntype{L}[1]{>{\raggedright\let\newline\\\arraybackslash\hspace{0pt}}m{#1}}
\newcolumntype{C}[1]{>{\centering\let\newline\\\arraybackslash\hspace{0pt}}m{#1}}
\newcolumntype{R}[1]{>{\raggedleft\let\newline\\\arraybackslash\hspace{0pt}}m{#1}}
\usepackage{booktabs}

\usepackage{paralist}

\usepackage[normalem]{ulem}
\usepackage{xfrac}

\usepackage[symbol]{footmisc}
\RequirePackage[normalem]{ulem} %DIF PREAMBLE
\RequirePackage{xcolor}
\usepackage{soul}
\usepackage{ulem}
\usepackage{censor}
\newlength\nextcharwidth
\makeatletter
\renewcommand\@cenword[1]{%
  \setlength{\nextcharwidth}{\widthof{#1}}%
  \censorrule{\nextcharwidth}%
  \kern -\nextcharwidth%
  #1}
\makeatother

\usepackage{algorithmicx}
\usepackage[ruled]{algorithm}
\usepackage{algpseudocode}
\usepackage{orcidlink}

\graphicspath{{figures/}}
\DeclareGraphicsExtensions{.pdf,.jpeg,.png,.eps}
\begin{document}
\title{Future Slot Prediction for Unsupervised Object Discovery in Surgical Video}
%
% \begin{comment}  %% Removed for anonymized MICCAI 2025 submission
\author{
%1{Guiqiu, Liao}
Guiqiu Liao\inst{1}\orcidlink{0000-0003-0921-178X} \and
%2{Matjaz, Jogan}
Matjaz Jogan\inst{1}\orcidlink{0000-0003-3771-3146} \and
%{Marcel, Hussing}
Marcel Hussing\inst{2} 
\and
%{Edward, Zhang}
Edward Zhang\inst{1,2} 
\and
%{Eric, Eaton}
\mbox{Eric Eaton\inst{2}} 
\and
%{Daniel, Hashimoto}
Daniel A. Hashimoto\inst{1,2}\orcidlink{0000-0003-4725-3104} 
}
\authorrunning{G. Liao et al.}
\institute{PCASO Laboratory, Department of Surgery,
University of Pennsylvania \and
Department of Computer and Information Science,
University of Pennsylvania
\email{Guiqiu.Liao@pennmedicine.upenn.com} }

% \end{comment}

% \author{Anonymized Authors}  %% Added for anonymized MICCAI 2025 submission
% \authorrunning{Anonymized Author et al.}
% \institute{Anonymized Affiliations \\
%     \email{email@anonymized.com}}

\maketitle              % typeset the header of the contribution
\begin{abstract}
\setcounter{footnote}{0}
Object-centric slot attention is an emerging paradigm for unsupervised learning of structured, interpretable object-centric representations (slots). This enables effective reasoning about objects and events at a low computational cost and is thus applicable to critical healthcare applications, such as real-time interpretation of surgical video. The heterogeneous scenes in real-world applications like surgery are, however, difficult to parse into a meaningful set of slots. Current approaches with an adaptive slot count perform well on images, but their performance on surgical videos is low. To address this
challenge, we propose a dynamic temporal slot transformer (DTST) module that is trained both for temporal reasoning and for predicting the optimal future slot initialization. The model achieves state-of-the-art performance on multiple surgical databases, demonstrating that unsupervised object-centric methods can be applied to real-world data and become part of the common arsenal in healthcare applications. {The code and models are publicly available at: {\color{blue} https://github.com/PCASOlab/Xslot}}.
\footnotetext{28th International Conference on Medical Image Computing
and Computer Assisted Intervention (MICCAI 2025)}

% Authors must provide keywords and are not allowed to remove this Keyword section.

\end{abstract}

\section{Introduction} % 1 page and a half max
\label{sec:intro}
Unsupervised object-centric learning seeks to bind features into modular latent representations of objects in visual scenes. Building on the foundations of variational auto-encoders \cite{kingma2013auto}, disentangled representation learning \cite{higgins2017beta}, iterative attention mechanisms \cite{sabour2017dynamic} and contrastive learning \cite{kipf2019contrastive}, \gls{sa} emerged as a dominant paradigm for object-centric learning \cite{locatello2020object,seitzer2022bridging,fan2024adaptive,mansouri2023object,jiang2023object,wu2023slotdiffusion}. In \gls{sa}, an attention mechanism is used to aggregate image pixels or features into slots, iteratively refining a representation optimized to map pixels to objects as in a set prediction problem using a reconstruction objective. 

In the temporal domain, recent efforts added object dynamics to self-su\-per\-vi\-sed objectives \cite{kipf2021conditional,singh2022simple,elsayed2022savi++,singh2024parallelized}. However, the permutation-invariant nature of set prediction in \gls{sa} poses challenges for the temporal consistency of slot allocations.  Surgical videos are a domain where these methods typically fail due to the non-uniform motion and intermittent visibility of instruments and anatomical structures. Recurrent architectures exploiting temporal feature similarity \cite{kipf2021conditional,singh2022simple,zadaianchuk2024object} and methods based on optical flow~\cite{kipf2021conditional} capture only limited, short-term causal interactions, and tend to drift over long sequences, struggling to capture long-range dependencies. Methods that batch process entire videos \cite{singh2024parallelized} increase computational demands. Slot-BERT \cite{liao2025slot} adopts the pre-training paradigm of large language models (LLMs), treating slot embeddings as word tokens to improve temporal coherence. All the aforementioned methods use a fixed number of slots per dataset, which can lead to over- or under-decomposition of the scene. Adaptive slot attention \cite{fan2024adaptive} improves object discovery by dynamically adjusting the number of slots to account for scene variability, yet it is ineffective on complex, rapidly changing surgical videos.

To address this, we propose a new architecture with a dynamic slot initialization module that predicts the appropriate number and allocation of slots in future frames. Inspired by next-token prediction in large language models, the \gls{dtst} processes sequences of slot embeddings and learns to predict future slots via a masked auto-encoding objective, effectively performing bidirectional temporal reasoning. Importantly, future slot prediction is jointly learned both as an initial slot allocation policy and as a post-refinement mechanism, further ensuring consistency and stability. Our model can operate on larger context window with minimal computational overhead by predicting slots in a latent space.
  With a two-stage training objective we achieve convergence and robustness across varying scenes. We validate our approach on multiple surgical video datasets, achieving state-of-the-art performance in unsupervised segmentation and localization of surgical instruments.

\section{Methods} % 2 pages
\label{sec:methods}

\begin{figure}[!t]
\includegraphics[width=0.99\textwidth]{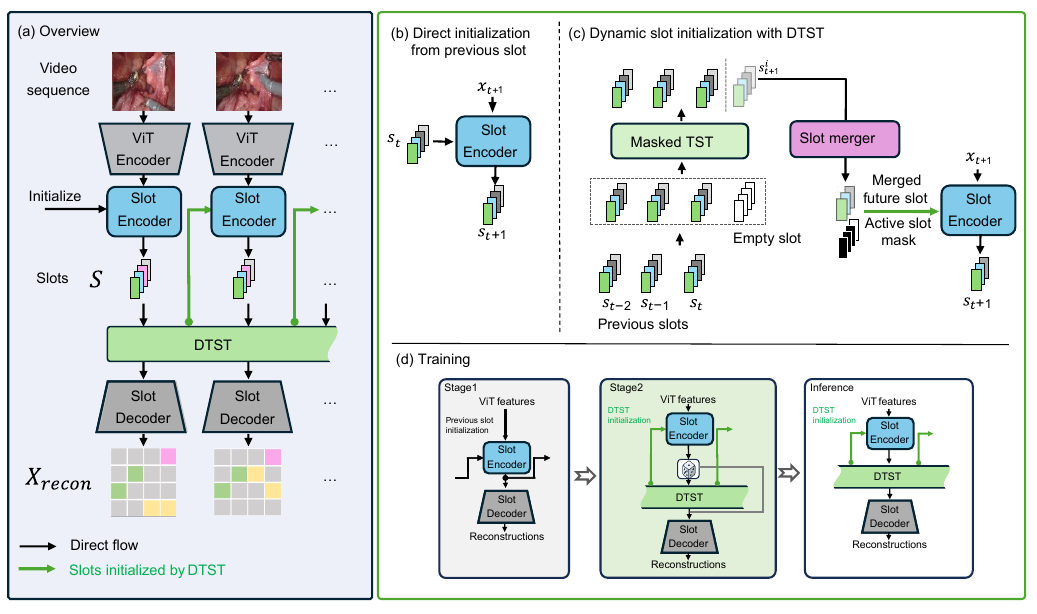}
\centering
\caption{(a) An overview of the proposed approach using DTST. (b) Direct initialization of slots at time $t+1$ from slot predictions at time $t$. (c) Dynamic slot initialization using DTST. (d) The overall training pipeline.
} 
\label{fig_method}
\end{figure}

Fig. \ref{fig_method}a shows an overview of our approach. The model processes videos of arbitrary length and iteratively operates on a buffered latent embedding of length \textit{T}, which is also the length of the attention window (e.g. temporal context window). A frame sequence  $\{I_t\}_{t=1}^T\in {\mathbb{R}}^{W\times H \times C \times T}$, where each $I_t \in {\mathbb{R}}^{W\times H\times C}$ represents a frame at time step $t$, is encoded to obtain features $X\in {\mathbb{R}}^{N \times D_{\textit{feature}}\times T}$. Following a recurrent iterative attention step we obtain the slots representation  $S = \{s_t\}_{t=1}^T\in {\mathbb{R}}^{K\times D_{\textit{slot}} \times T}$, where $s_t$ are the latent space slots that embed objectness of image $I_t$.  Slots $S$ are then fed to \gls{dtst} encoder and a merger module that aggregates the between-slots information and calculate $S_{\textit{merged}}$, allocating the redundant slots to new objects that entered the scene, removing the slots of objects that exit the scene, and merging multiple slots of different parts of the same object. 
A slot decoder then recurrently maps each element $s_t$ of $S_{\textit{merged}}$ back to the video encoding space  $X_{\textit{recon}} \in {\mathbb{R}}^{N\times D_{\textit{feature}}\times T}$. Simultaneously, the object segmentation masks for each slot are reconstructed. The only training objective is to minimize 
\begin{equation}
% \nonumber
\mathcal{L}_{\text{recon}} = \|X_{\text{recon}} - X\|^2 \enspace 
\end{equation} 
which measures the distance between the reconstructed and original features $X$ \cite{seitzer2022bridging}. Feature-level reconstruction is more robust to variations in scene appearance than image reconstruction.
  
%%%%%%%%%%%%%%%%%%%%%%%%%%%%%%%%%%%%%%%%%%%%%%%%%%%%%%%%%%%%%%%%%%%%%%%%%%%%%%%%%%%%%
\subsection{Slot encoder}
\label{subsec:methods_encoding_scheme}
 
The slot encoder $f_{\mathrm{SA}}$ \cite{locatello2020object} groups image features $x\in {\mathbb{R}}^{N \times D_{\textit{feature}}}$ into $K$ spatial groupings:  
\begin{equation} 
%\nonumber
s := f_{\mathrm{SA}}(x, s^{\,i}) 
\end{equation} 
  where $s^{\,i} \in {\mathbb{R}}^{K\times d_{\textit{slot}}} $ is the initialization slot. $K$ is a predefined constant to determine the size of attention query size.  

\subsection{Slot decoder}
\gls{mlp} broadcast decoders efficiently decode features and slot masks \cite{watters2019spatial,seitzer2022bridging}. Each of the $K$ slots is broadcast to match the number of spatial patches $N$, resulting in $N$ tokens for each slot. A learnable positional encoding is added to each token; these are then processed independently using a shared \gls{mlp} to output reconstructed features $\hat{x}_k$ and associated alpha masks $\alpha_k$ indicating the slot's attentive region. The final reconstruction $x \in \mathbb{R}^{N \times D_{\text{feature}}}$ is obtained by a weighted sum:
\begin{equation}
%\nonumber
x = \sum_{k=1}^K \hat{x}_k \odot m_k\ , \quad m_k = \text{softmax}_k(\alpha_k)
\end{equation} 
where $\odot$ denotes element-wise multiplication. This simple design is very efficient: as the MLP is shared across slots and positions, $m_k$ is directly inferred as a set of non-overlapping object segmentation masks. 

\subsection{Pre-training}
%The query $q$ in slot attention is a function of the slots $s^i$, and are iteratively refined over several iterations. 
We pre-train the encoder and decoder using a simple slot assignment via a RNN-like computation, as illustrated in Fig. \ref{fig_method} b. The first frame's slots $s_0^{\,i=0}$ are initialized randomly by sampling from a standard Gaussian. Slots $s_t^{\,i=0}$ =$\hat{s}_{t-1}$ for $t$-th frame are initialized by using the slot prediction from the prior frame, encouraging stability of slot identity across frames. 

\subsection{Dynamic future slot prediction}
%\begin{figure}[!t]
%\includegraphics[width=0.7\textwidth]{figures/training.pdf}
%\centering
%\caption{ Training pipeline. The stochastic DTST drop is introduced for the second %stage.
%}
% 
% \label{fig_tr}
% \end{figure}
After pretraining,  a Dynamic Temporal Slot Transformer (\gls{dtst}) and a slot merger are enabled to facilitate slot interactions across frames. \gls{dtst} is based on the same transformer architecture as in language models \cite{vaswani2017attention,radford2019language}, replacing word tokens with the video slot embedding vectors $S \in \mathbb{R}^{K\times d_{slot} \times T}$. Importantly, temporal positional embeddings and random masking-out of slots also helps predict missing masked slots, which will be crucial for dynamic slot initialization.

\noindent\textbf{Slot merger.}
Unlike previous methods including Slot-BERT \cite{liao2025slot}, which use all available slots for reconstruction, we employ a slot merger that resolves the problem of over-grouping or under-grouping given a fixed slot count $K$. Inspired by slot merging using clustering  \cite{aydemir2023self}, we design a method that works directly on slot similarity. Given a sequence of slot representations $S \in \mathbb{R}^{K \times d_{\text{slot}} \times T}$,
we compute the cosine similarity between each pair of slots \(s_i, s_j \in \mathbb{R}^{d_{\text{slot}}}\) as
\begin{equation}
\text{sim}(s_i, s_j) = \frac{s_i \cdot s_j}{\|s_i\|\, \|s_j\| + \epsilon}\, 
\end{equation}
Thresholding similarity scores results in a binary merge mask identifying similar slots that are then averaged, giving an updated slot representation 
$\tilde{S}_{merge}$. In addition, a binary mask $
m_s \in \{0,1\}^{T \times K}$
indicates which slots remain valid after merging. 
% In training, the slot merger is used with a droppath to ensure that \gls{dtst} and the decoder learn to handle varying slot numbers. 
Our slot-merger module is trained within the unsupervised pipeline with a drop-path mechanism, enabling the \gls{dtst} and the decoder to adapt to varying slot counts. In contrast to Slot-Roll-Out \cite{zhao2023object} which requires sparse supervision, our method performs unsupervised merging of redundant slots and integrates object parts solely based on appearance and dynamics. 

\noindent\textbf{Dynamic temporal slot transformer.} \gls{dtst} is used for bidirectional temporal reasoning using random masking in training. As illustrated by Fig. \ref{fig_method} c, the \gls{dtst} module is also used to initialize the recurrent estimation of future slots. To do this, we reformulate the module as next slot prediction by appending empty slots, initialized as zero vectors, to the most recent slot buffer. The \gls{dtst} module then predicts the future slot using current slot history and a single slot mask for last position. We also apply the slot merger to these predictions. 
 This design offers two key advantages: 1) Existing slots are initialized more accurately, improving the tracking of moving objects; and 2) Adjusting the number of active slots with the slot merger prior to the competitive \gls{sa} mechanism allows the slot encoder to form more optimal groupings. 
 
 To reinforce model convergence, as shown in Fig. \ref{fig_method} d, we allow the model to stochastically bypass the dynamic \gls{dtst} before the decoding step while keeping it active in slot initialization. This design allows the module to specialize both as a next-slot predictor and as a temporal reasoner, aiding the decoder in reconstructing both refined (without the slot merger) and original slots, resulting in a more robust training framework. The resulting training framework gains robustness by co-learning the context and the future predictions; aligning thus with modern LLM practices \cite{radford2019language} that are more effective than in-context training of architectures like Slot-BERT \cite{liao2025slot}.

%%%%%%%%%%%%%%%%%%%%%%%%%%%%%%%%%%%%%%%%%%%%%%%%%%%%%%%%%%%%%%%%%%%%%%%%%%%%%%%%%%%%%

\section{Experiments and results} % 3 1/2 - 4 pages max
\label{sec:experiments_results}

\subsection{Datasets and metrics}
\label{sec_dataset}
We evaluate performance on four surgical video datasets spanning three surgery types: (1) an Abdominal surgery dataset based on the public MICCAI 2022 SurgToolLoc Challenge Data \cite{zia2023surgical} that includes 24,542 training clips and 100 testing clips from animal, phantom, and simulator surgeries, downsampled to 1 FPS and with 13 instrument classes; (2) a proprietary Thoracic Robotic Surgery dataset that includes 1,730 training clips and 264 testing clips annotated with instruments sampled at 6 FPS from 40 robotic lung lobectomy videos; (3) a  cholecystectomy dataset  based on Cholec80 \cite{twinanda2016endonet} that consists of 5,296 training and 100 testing 30-frame clips sampled at 1 FPS from cholecystectomy videos, where the testing clips are extracted from the CholecSeg8K \cite{hong2020cholecseg8k} semantic segmentation dataset; and (4) EndoVis 2017 dataset with 480 clips sampled at 1 FPS (five-frame sequences from abdominal surgery) that is used for zero-shot evaluation. We evaluate our approach by comparing the slot masks produced by the decoder with ground-truth annotations of instrument masks using video mean best overlap (mBO-V) \cite{pont2016multiscale}, frame mean best overlap (mBO-F), mean best Hausdorff Distance (mBHD), and foreground adjusted rand index (FG-ARI) \cite{greff2019iodine}. We also use CorLoc \cite{fan2024adaptive} to evaluate the instrument bounding box localization accuracy. mBO-F, mBHD, FG-ARI and CorLoc measure overlap with ground truth while mBO-V additionally evaluates the temporal consistency of instrument segmentation.
% \begin{tablenotes}
% \small
% \item[1] *When testing on thoracic the frame length is  

\begin{table}[t!]
\caption{Comparison to state-of-the-art unsupervised object-centric learning methods on abdominal, thoracic and cholecystectomy datasets.}
\label{tab_sota}
\centering
\setlength\arrayrulewidth{0.9pt}
\setlength\doublerulesep{0.9pt} 
\setlength{\tabcolsep}{5pt} % Reduced column spacing

\resizebox{1.0\linewidth}{!}{%
\begin{tabular}{|c|c|ccccc|}
\hline
\multicolumn{1}{|l|}{} & \multicolumn{1}{l|}{Method} & mBO-V & mBO-F & mBHD (↓) & FG-ARI & CorLoc \\ \hline
\multirow{9}{*}{Abdominal}   
 & SAVi\cite{kipf2021conditional} & 29.4±0.2 & 33.2±0.1 & 81.72±0.99 & 36.6±0.2 & 40.0±0.5 \\  
 & STEVE\cite{singh2022simple} & 27.9±0.2 & 31.5±0.1 & 139.93±0.62 & 34.3±0.1 & 17.0±0.4 \\  
 & Slot-Diffusion\cite{wu2023slotdiffusion} & 37.5±0.1 & 42.2±0.1 & 70.54±0.25 & 46.3±0.0 & 42.0±0.2 \\  
 & Video-Saur\cite{zadaianchuk2024object} & 46.3±0.4 & 50.1±0.4 & 53.90±1.28 & 55.1±0.5 & 60.0±1.9 \\  
 & Adaslot\cite{fan2024adaptive} & 38.2±0.9 & 41.0±0.6 & 66.58±0.88 & 47.9±0.5 & 55.1±0.4 \\  
 & Adaslot-ext & 36.5±0.5 & 39.9±0.8 & 65.03±1.86 & 46.9±0.9 & 55.7±0.4 \\  
 & Slot-BERT\cite{liao2025slot} & 48.9±0.2 & 52.8±0.2 & 43.40±0.53 & 58.2±0.3 & 70.7±0.8 \\  
 & Ours & \textbf{50.8±0.2} & \textbf{54.1±0.5} & \textbf{40.21±0.66} & \textbf{62.8±0.3} & \textbf{71.7±1.0} \\\hline  
\multirow{9}{*}{Thoracic}  
 & SAVi\cite{kipf2021conditional} & 26.5±0.0 & 30.7±0.1 & 119.76±0.17 & 23.1±0.0 & 27.3±0.7 \\  
 & STEVE\cite{singh2022simple} & 28.9±0.0 & 33.0±0.0 & 135.41±0.54 & 26.5±0.0 & 24.6±0.3 \\  
 & Slot-Diffusion\cite{wu2023slotdiffusion} & 31.1±0.1 & 39.9±0.0 & 84.56±0.15 & 31.4±0.1 & 31.7±0.5 \\  
 & Video-Saur\cite{zadaianchuk2024object} & 21.9±0.1 & 15.7±0.1 & 139.46±0.23 & 11.9±0.1 & 13.0±0.1 \\  
 & Adaslot\cite{fan2024adaptive} & 23.0±0.5 & 28.4±0.6 & 116.15±1.16 & 21.7±0.2 & 25.1±1.6 \\  
 & Adaslot-ext & 33.7±0.3 & 46.2±0.2 & 80.51±0.56 & 37.6±0.3 & 44.5±0.3 \\  
 & Slot-BERT\cite{liao2025slot} & 34.0±0.1 & 40.5±0.1 & 92.37±0.43 & 33.8±0.1 & 34.8±0.7 \\  
 & Ours & \textbf{38.2±0.1} & \textbf{52.2±0.1} & \textbf{68.17±0.65} & \textbf{41.6±0.1} & \textbf{54.5±0.3} \\\hline  
\multirow{9}{*}{Cholecystectomy}  
 & SAVi\cite{kipf2021conditional} & 24.8±0.0 & 29.4±0.0 & 96.51±0.21 & 20.1±0.0 & 26.7±0.3 \\  
 & STEVE\cite{singh2022simple} & 26.2±0.1 & 32.3±0.0 & 101.55±0.19 & 24.9±0.1 & 30.1±0.4 \\  
 & Slot-Diffusion\cite{wu2023slotdiffusion} & 27.7±0.2 & 34.6±0.1 & 94.70±0.37 & 26.3±0.2 & 38.5±0.7 \\  
 & Video-Saur\cite{zadaianchuk2024object} & 30.1±0.3 & 43.1±0.1 & 75.82±0.24 & 34.2±0.1 & 34.1±0.8 \\  
 & Adaslot\cite{fan2024adaptive} & 17.5±0.9 & 17.8±0.3 & 95.76±1.22 & 22.4±0.8 & 19.0±0.2 \\  
 & Adaslot-ext & 31.6±0.5 & 31.1±0.3 & 60.51±0.39 & 42.4±0.1 & 33.7±1.0 \\  
 & Slot-BERT\cite{liao2025slot} & 28.8±0.3 & 27.8±0.4 & 64.82±0.90 & 37.0±0.6 & 35.3±2.0 \\  
 & Ours & \textbf{33.2±0.8} & \textbf{31.9±0.5} & \textbf{57.32±0.88} & \textbf{43.1±0.9} & \textbf{41.2±0.4} \\\hline  
\end{tabular}
}
% \begin{tablenotes}
% \small
% \item[1] *When testing on thoracic the frame length is increased in comparison to Endovis and Cholec
% \end{tablenotes}
\end{table}

 \subsection{Results}

\begin{figure}[t!]
    \centerline{
    \includegraphics[width=0.99\linewidth]{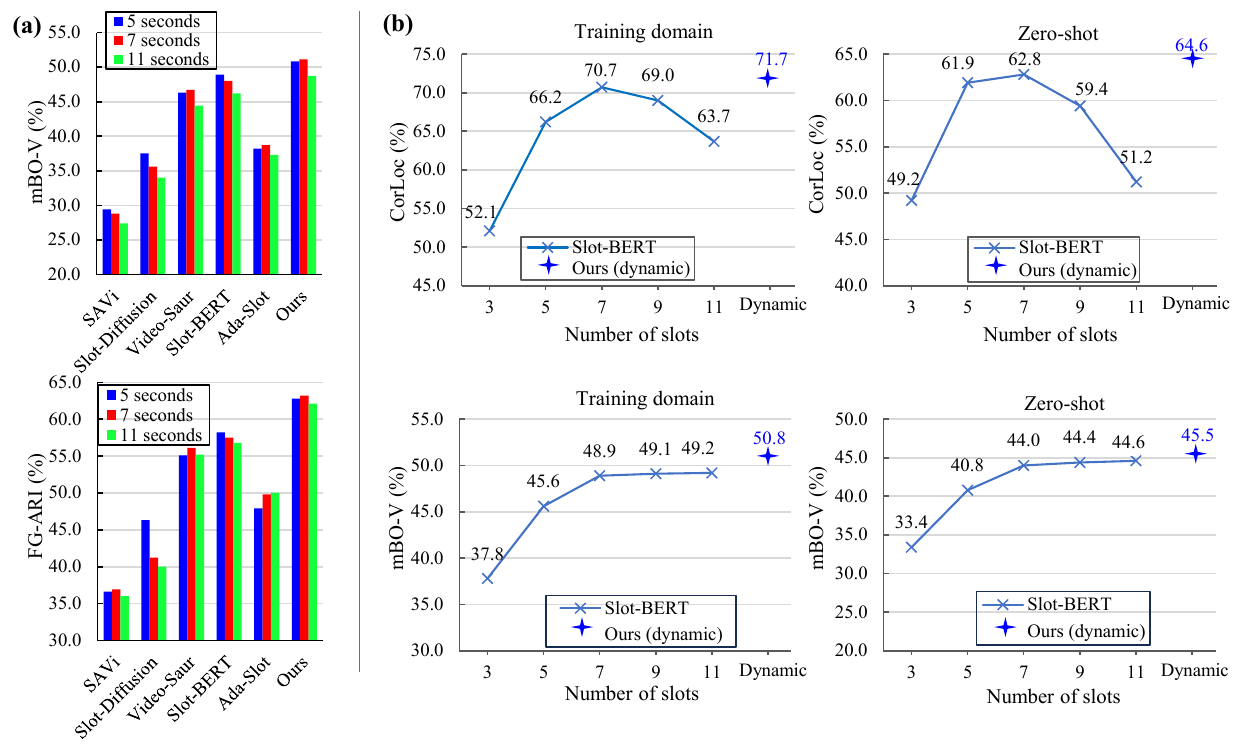}}
    \caption{ (a) Quantitative comparison to five state-of-the-art methods on videos of lengths 5, 7 and 11 seconds. (b) Comparison of our dynamic future slot prediction to Slot-BERT with different static slot counts in different domains.} 
    \label{fig_length}
\end{figure}

\noindent\textbf{Comparison to state-of-the-art.} We compare our method to other unsupervised object representation learning methods for video including SAVi \cite{kipf2021conditional}, STEVE \cite{singh2022simple}, Slot-Diffusion \cite{wu2023slotdiffusion}, Video-Saur \cite{zadaianchuk2024object}, and Slot-BERT \cite{liao2025slot}. The only slot attention method in the literature that implements  dynamic slot allocation is AdaSlot \cite{fan2024adaptive} which was developed for images. We thus reimplement AdaSlot for videos using its original slot prediction layer and an extended prediction layer (AdaSlot-ext) to account for the increased complexity of surgical scenes.
As shown in Table \ref{tab_sota}, our method achieves superior performance across all datasets. The improvement is particularly large for the thoracic dataset, which is also the smallest, suggesting that our method requires significantly less training data. 
 On the abdominal dataset, the CorLoc score of 71.7 approaches the level of performance of supervised methods. Statistical significance of this
improvement is confirmed by one tailed t-test ($p$-values < 0.01 with Bonferroni correction).

 Results in Table \ref{tab_sota} are calculated on five-second clips. Fig \ref{fig_length}~(a) demonstrates that our method is consistently better on longer clips (7 and 11 seconds) measured by mBO-V and FG-ARI. Longer clips (29 seconds each) of object discovery and segmentation are available in the Supplemental Material.
 
\noindent\textbf {Comparison with tuned slot counts.}
Fig. \ref{fig_length}~(b) illustrates our method’s performance in localization (CorLoc) and video segmentation (mBO-V) compared to the second-best method, Slot-BERT, across different slot counts. On the training domain (Abdominal), Slot-BERT's  localization accuracy declines as the number of slots increases despite a slight improvement in video segmentation. In contrast, our method outperforms slot-BERT in both localization and video-level segmentation, all  without a fixed slot count. Additionally, our method leads in zero-shot transfer to the EndoVis dataset, achieving the highest CorLoc (64.6) and mBO-V (45.5), thus demonstrating a superior generalization.

\begin{figure}[t!]
    \centerline{
    \includegraphics[width=1.0\linewidth]{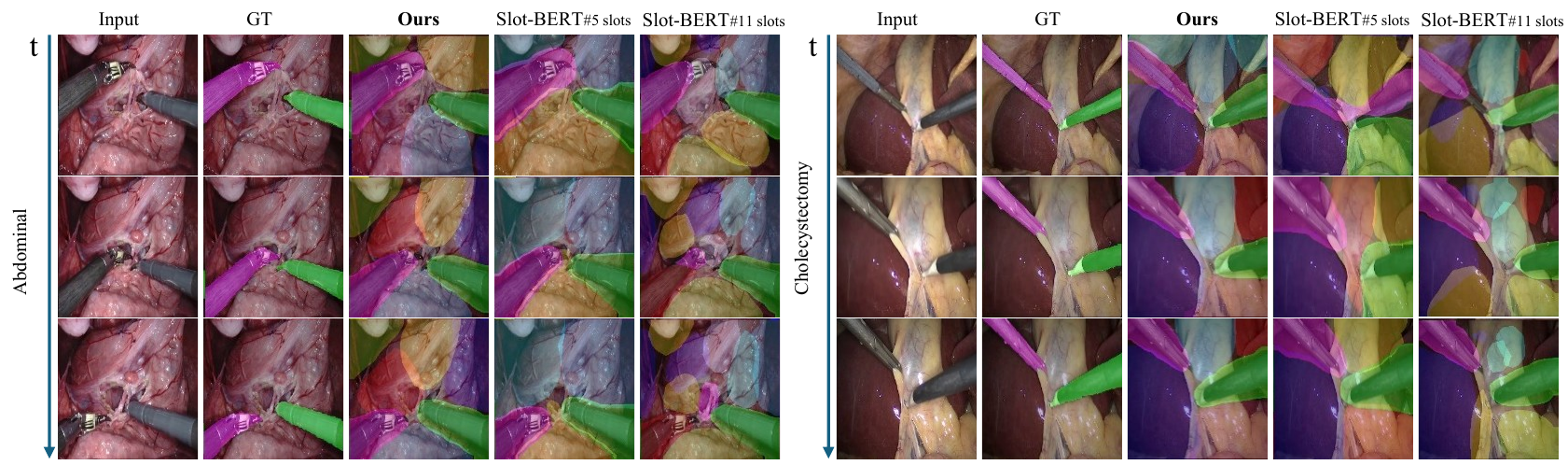}}
    \caption{Qualitative results of unsupervised segmentation using dynamic future slot prediction for abdominal and cholecystectomy surgery videos, compared to Slot-BERT with a fixed number of slots (5 and 11 slots). Ground Truth (GT) on instruments is provided as reference.} 
    \label{fig_seg}
\end{figure}

Fig \ref{fig_seg} further demonstrates the advantage of dynamic slot allocation and advanced temporal reasoning: A fixed slot count used by Slot-BERT tends to over-segment (11 slots) or under-segment (5 slots), while our method finds a balanced slot count that better delineates instruments and tissues.

\noindent\textbf{Ablation study.} Table \ref{tab_abla} shows the contributions of different model components: \gls{dtst}, slot merger (Merger), and their application to future/next-slot prediction (X-slot) to performance in same domain for a short (5 sec) and long (11 sec) sequence, and in a zero-shot transfer to the EndoVis dataset. Individually, \gls{dtst} improves temporal consistency (e.g., mBO-V increases from 45.6 to 46.2 in short sequences), while Merger significantly enhances object localization (e.g., CorLoc improves from 60.1 to 70.0 in long sequences). Next slot prediction further boosts performance, particularly in zero-shot settings (CorLoc increases from 60.9 to 64.6). The full model (\gls{dtst} + Merger + X-slot) achieves the best overall performance, with notable improvements in temporal consistency as reflected by mBO-V. These results highlight the complementary strengths of these components in achieving robust object-centric learning.

\begin{table}[t!]
\caption{Ablation study. }
\label{tab_abla}
\centering
\setlength\arrayrulewidth{0.9pt}
\setlength\doublerulesep{0.9pt} 

\resizebox{1.0\linewidth}{!}{%
\begin{tabular}{|ccc|cccc|cccc|cccc|}
\hline
 &  &  & \multicolumn{8}{c|}{Same domain} & \multicolumn{4}{c|}{\multirow{2}{*}{Zero-shot}} \\ \cline{4-11}
 &  &  & \multicolumn{4}{c|}{Short} & \multicolumn{4}{c|}{Long} & \multicolumn{4}{c|}{} \\ \hline
\multicolumn{1}{|c}{\textbf{DTST}} & \multicolumn{1}{c}{\textbf{Merger}} & \multicolumn{1}{c|}{\textbf{X-slot}} & mBO-V & mBHD(↓) & FG-ARI & CorLoc & mBO-V & mBHD(↓) & FG-ARI & CorLoc & mBO-V & mBHD(↓) & FG-ARI & CorLoc \\ \hline
 &  &  & 45.6 & 44.5 & 59.7 & 61.7 & 42.5 & 44.3 & 58.5 & 60.1 & 39.4 & 53.5 & 53.1 & 48.9 \\
\checkmark &  &  & 46.2 & 44.6 & 60.1 & 62.2 & 43.7 & 44.0 & 59.9 & 60.2 & 40.6 & 51.7 & 55.1 & 49.0 \\
 & \checkmark &  & 45.4 & 41.8 & 62.8 & 70.8 & 45.1 & 42.7 & 62.5 & 70.0 & 43.4 & 49.5 & 57.2 & 59.1 \\
\checkmark &  & \checkmark & 49.5 & \textbf{40.0} & \textbf{63.2} & 66.3 & 45.9 & 40.4 & 61.4 & 63.9 & 43.4 & 47.0 & 56.5 & 55.8 \\
\checkmark & \checkmark &  & 49.2 & 41.0 & {63.1} & 71.7 & 46.5 & 41.2 & \textbf{63.1} & 72.0 & 44.3 & 48.4 & \textbf{58.0} & 60.9 \\
\checkmark & \checkmark & \checkmark & \textbf{50.8} & {40.2} & 62.8 & 71.7 & \textbf{48.7} & \textbf{39.5} & 62.1 & \textbf{73.4} & \textbf{45.5} & \textbf{44.8} & 57.9 & \textbf{64.6} \\ \hline
\end{tabular}
}
\end{table}

\noindent\textbf{Effect of cosine-similarity threshold.}
We tested robustness to the slot similarity threshold (Table \ref{tab_thre}). 
 Both mBO-V and FG-ARI improve consistently with higher thresholds, peaking at 0.99, suggesting stricter similarity refines slot assignments. Meanwhile, CorLoc peaks at 0.85 (73.0) and declines for stricter thresholds, as higher thresholds reduce the likelihood of merging multiple slots belonging to the same object.
 While there's a trade-off between localization and segmentation, thresholds above 0.80 yield stable performance, showing threshold selection is more permissive than slot count selection. Our method also allows a different number of active slots in each frame which is a clear advantage over fixed slot count approaches.

\noindent\textbf{Computational efficiency.} The forward time per frame on a single NVIDIA RTX A6000 GPU is 5.6 ms, supporting real-time downstream tasks. While slot merging and future slot prediction add 3.9 ms overhead compared to Slot-BERT (1.7 ms), this remains minor. Even with a 20$\times$ larger context window, latency stays under 100 ms thanks to latent-space temporal reasoning.

\begin{table}[h!]
\caption{Influence of slot similarity threshold.}
\label{tab_thre}
\centering
\setlength\arrayrulewidth{0.9pt}
\setlength\doublerulesep{0.9pt} 
\setlength{\tabcolsep}{7pt} % Reduced column spacing
\resizebox{0.9\linewidth}{!}{%
\scriptsize % This ensures the text remains small inside the resized box
\begin{tabular}{|c|cccccc|} 
 \hline
Matrics &    0.70  & 0.80 & 0.85 & 0.90 & 0.95 & 0.99   \\  \hline
CorLoc & 58.6±1.5 & 72.3±1.1 & \textbf{73.0±0.9} & 71.7±1.0 & 70.3±0.6 & 70.2±0.2   \\
mBO-V  & 38.9±0.7 & 48.9±0.4 & 49.4±0.2 & 50.8±0.2 & 50.9±0.2 & \textbf{51.6±0.5}   \\ 
FG-ARI & 42.8±1.1 & 59.9±0.6 & 61.0±0.6 & 62.8±0.3 & 62.8±0.7 & \textbf{63.1±0.6}  \\ \hline

\end{tabular}
}
\end{table}

\section{Conclusion}
\label{sec:conclusion}

We proposed a novel approach for temporal reasoning and future slot initialization in Slot Attention (SA) that improves object discovery in video and shows promising results on the complex task of surgical video understanding.  The SA model is coupled with our  Dynamic Temporal Slot Transformer (DTST) and a slot merger to extract object representations and predict object masks. We train DTST stochastically, making it efficient for both temporal reasoning in post-attention and for next slot initialization. Experiments show that our approach performs robustly across multiple surgical video domains and video lengths, highlighting its adaptability and generalization, including in zero-shot transfer.

Slot attention jointly embeds appearance and location in the latent space, allowing slot tracking even when objects are partially or temporarily occluded. However, extreme overlaps of objects might be still difficult to disentangle. E.g., instruments with longer, uniform shafts will be correctly outlined after initial occlusion (due to location bias), but might split beyond the context window (i.e., >11 sec). Exploring solutions with a longer context window and a specialized training set is part of our future work.

The state-of-the-art performance demonstrated on multiple surgical databases already opens possibilities for downstream applications that require interpretable surgical video understanding with low computing demands. A larger training dataset drastically improves performance (see results for the abdominal dataset). Thus, incorporating more diverse training data could pave the way for broader applications in healthcare and beyond.
 % which allow the model to learn meaningful interpretable representations, P

\subsubsection{Acknowledgements.} 

This work was supported by the Linda Pechenik Montague Investigator Award and the American Surgical Association Foundation Fellowship.
% Anonymous acknowledgments.

\subsubsection{Disclosure of interests.} The authors have no competing interests to declare that
are relevant to the content of this article.

% Anonymous version:
% \subsubsection{Acknowledgements.} 
% ************************************************\\**********************************************************************\\**********************************************************************\\**********************************************************************\\**********************************************************************

% Anonymous version 2:
% \subsubsection{Acknowledgements.} Anonymous funding\\

%%%%%%%% TEMPORARY PAGE BREAK TO ASSESS THE REAL LENGTH OF THE SECTIONS AND THE REFERENCES %%%%%%%%

% \pagebreak % use page break to check
\bibliography{bibliography/refs_miccai.bib}
\bibliographystyle{splncs04}

\end{document}